%% file: 0._Main.tex
\documentclass[runningheads]{llncs}
\usepackage[T1]{fontenc}
\usepackage{graphicx}
\usepackage{amsmath}
\usepackage{amssymb}
\usepackage{booktabs}
\usepackage{algorithm}
\usepackage{algpseudocode}
\usepackage{multirow} 
\usepackage{orcidlink}
\usepackage{setspace}
\usepackage{authblk}

\begin{document}
\title{Zero-Shot Industrial Anomaly Segmentation \\ with Image-Aware Prompt Generation}
\titlerunning{ZSAS with Image-Aware Prompt Generation}

\author{SoYoung Park\inst{1}\textsuperscript{*}\orcidlink{0009-0000-7755-0041}\and 
Hyewon Lee\inst{1}\textsuperscript{*}\orcidlink{0009-0003-9573-2916}\and
Mingyu Choi\inst{1}\orcidlink{0009-0006-2909-150X} \and
Seunghoon Han\inst{1}\orcidlink{0009-0008-3393-4797} \and
Jong-Ryul Lee\inst{1}\textsuperscript{†}\orcidlink{0000-0003-0774-3619} \and
Sungsu Lim\inst{1}\textsuperscript{†‡}\orcidlink{0000-0001-5924-3398} \and 
Tae-Ho Kim\inst{2}\textsuperscript{†}\orcidlink{0000-0003-4522-5388}
}
\authorrunning{S.Y. Park et al.}

\institute{Department of Computer Science and Engineering, \\
Chungnam National University, Daejeon, South Korea\\
\email{{sypark1452,noweyh927}@o.cnu.ac.kr, {mingyu,tmdgns129}@g.cnu.ac.kr, {jongryul.lee,sungsu}@cnu.ac.kr}\\ 
\and Nota Inc., Seoul, South Korea \\
\email{thkim@nota.ai}\\
}

\maketitle

\def\thefootnote{*}\footnotetext{Equal contribution.}
\def\thefootnote{†}\footnotetext{Corresponding authors.}
\def\thefootnote{‡}\footnotetext{This work was done when the author was a visiting researcher at Nota AI.}

\input{0.abstract}

\input{1.introduction}

\input{2.related_work}

\input{3.methodology}

\input{4.experiments}

\input{5.conclusions}

\begin{credits}
\subsubsection{\ackname} This work was supported by the Institute of Information \& Communications Technology Planning \& Evaluation (IITP) grant funded by the Korea government (MSIT) (No. RS-2022-00155857, Artificial Intelligence Convergence Innovation Human Resources Development (Chungnam National University)) and by the Technology Innovation Program (RS-2024-00468747, Development of AI and Lightweight Technology for Embedding Multisensory Intelligence Modules) funded by the Ministry of Trade Industry \& Energy (MOTIE, Korea).

\subsubsection{\discintname}
The authors have no competing interests to declare that are
relevant to the content of this article.
\end{credits}

\bibliographystyle{splncs04}
\bibliography{egbib}

\end{document}

%% file: 0.abstract.tex
\begin{abstract}
Anomaly segmentation is essential for industrial quality, maintenance, and stability. Existing text-guided zero-shot anomaly segmentation models are effective but rely on fixed prompts, limiting adaptability in diverse industrial scenarios. This highlights the need for flexible, context-aware prompting strategies. We propose Image-Aware Prompt Anomaly Segmentation (IAP-AS), which enhances anomaly segmentation by generating dynamic, context-aware prompts using an image tagging model and a large language model (LLM). IAP-AS extracts object attributes from images to generate context-aware prompts, improving adaptability and generalization in dynamic and unstructured industrial environments. In our experiments, IAP-AS improves the F1-max metric by up to 10\%, demonstrating superior adaptability and generalization. It provides a scalable solution for anomaly segmentation across industries.

\keywords{Zero-shot anomaly segmentation (ZSAS) \and Large language model (LMM) \and Text prompts \and Image tagging model}
\end{abstract}

%% file: 1.introduction.tex
\section{Introduction} \label{sed:intro}
\begin{figure*}[t]
  \centering
  \includegraphics[width=0.8\linewidth]{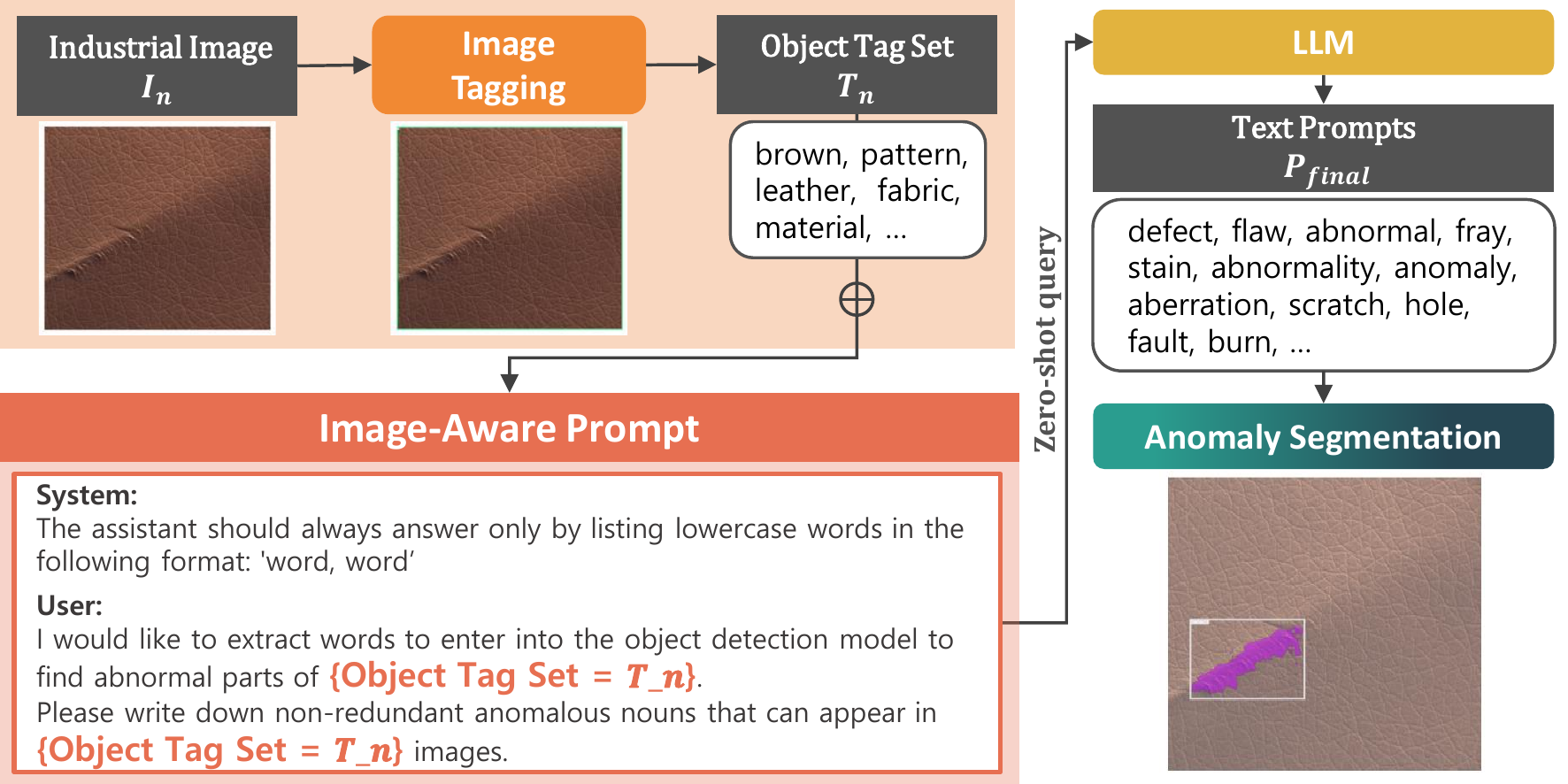}
  \caption{Example of the text prompt generation process for an industrial image, where object tags extracted from the image are combined with Image-Aware Prompt (IAP) and processed by an LLM to create context-aware prompts for anomaly segmentation.}
  \label{fig:Prompting-example}
\end{figure*}

Anomaly segmentation is vital for industrial tasks such as quality assurance, maintenance, and process stability~\cite{zipfel2023anomaly,bao2023miad}. However, challenges arise due to the diverse nature of defects and the difficulty in defining anomalies. 
Traditional models rely on large annotated datasets, limiting adaptability and increasing costs. 
Furthermore, the scarcity  of anomaly samples further restricts cross-domain generalization~\cite{as2}.
Recent advancements in Zero-Shot Anomaly Segmentation (ZSAS) have enabled effective generalization without anomaly samples. Among these approaches, text-guided ZSAS models have gained significant attention. These models leverage pre-trained large-scale models such as Segment Anything Model (SAM)~\cite{SAM} and Contrastive Language-Image Pre-training (CLIP)~\cite{CLIP} to recognize object attributes in images and align image features with semantic information using text prompts~\cite{text}. 
These models excel in zero-shot scenarios, making them ideal for cases with limited or no labeled anomaly data.

However, text-guided ZSAS models encounter challenges in real-world applications. CLIP-based models struggle with fine-grained anomaly segmentation, while SAM-based models require complex prompts and extensive post-processing, limiting their practicality in dynamic environments~\cite{cris,ClipSAM,SAA}. Furthermore, reliance on fixed prompts restricts cross-domain generalization. 
For instance, in an industrial setting, a fixed prompt such as "detect scratches" may fail when applied to products with varying surface materials, as the appearance of scratches differs significantly between metal and glass.
This highlights the need for adaptive, context-aware prompting mechanisms~\cite{clipad,prompt-eng}.

To address these limitations, we propose \textbf{Image-Aware Prompt Anomaly Segmentation (IAP-AS)}, which integrates an image tagging model~\cite{RAM} and an LLM~\cite{Llama} to generate context-aware prompts. Unlike fixed prompts, IAP-AS dynamically identifies objects and their attributes, enabling more accurate and flexible anomaly segmentation across industrial domains, as illustrated in Fig.~\ref{fig:Prompting-example}. By leveraging object detection~\cite{dino} and high-precision segmentation~\cite{SAM} models, it enhances adaptability and generalization without retraining.
This study introduces IAP-AS, a novel ZSAS approach that addresses the limitations of fixed prompts through image-aware prompts. Our key contributions include:
\begin{itemize}
    \item \textit{Dynamic Context-Aware Prompting}: IAP-AS generates adaptive text prompts based on image content, improving segmentation accuracy.
    \item \textit{Generalization Across Various Industrial Domains}: IAP-AS generalizes across industrial domains without retraining, ensuring robust performance.
    \item \textit{Reproducibility}: The proposed method is publicly available  at \url{https://github.com/sybeam27/IAP-ZSAS}, ensuring reproducibility.
\end{itemize}

%% file: 2.related_work.tex
\section{Related Work} \label{sec:relatedwork}
Zero-Shot Anomaly Segmentation (ZSAS) methods are classified into training-free and training-required approaches. Training-free methods~\cite{ClipSeg,WinCLIP,SAA,clipad} offer superior adaptability and efficiency and are further divided into CLIP-based and SAM-based models, each with distinct strengths and limitations.

CLIP-based methods, including WinCLIP~\cite{WinCLIP}, CLIPSeg~\cite{ClipSeg}, and SDP~\cite{clipad}, utilize global feature alignment to associate image features with text prompts, leveraging large-scale pre-trained models like CLIP~\cite{CLIP} for semantic matching. However, these methods have limitations. Their dependence on global features hinders the detection of fine-grained anomalies, often missing small or subtle deviations~\cite{clipad}. Additionally, models like WinCLIP employ a window-based encoding strategy to improve feature alignment, but this increases computational cost and reduces efficiency~\cite{cris}.
On the other hand, SAM-based methods, such as SAA+~\cite{SAA} and EVF-SAM~\cite{evf}, utilize SAM~\cite{SAM} to generate candidate segmentation masks using hybrid prompts, refining them through post-processing. Although this approach improves segmentation accuracy, it also introduces several challenges. The generation of multiple candidate masks often causes redundancy and over-segmentation, leading to unnecessary computations~\cite{clipad}. Moreover, the reliance on post-processing for mask refinement limits real-time applicability, while the need for prompt adjustments across datasets reduces cross-domain generalization~\cite{ClipSAM,clipad}.

Effective prompt design is crucial for ZSAS, as segmentation performance depends on prompt quality~\cite{clipad}. Existing approaches fall into three types: fixed, few-shot, and pre-defined prompts.
Many ZSAS methods utilize fixed prompts such as "abnormal" or "defect." While simple and intuitive, these prompts fail to account for domain-specific variations, leading to suboptimal performance on unseen datasets. Few-shot prompting techniques, on the other hand, evaluate multiple prompts on a small set of samples from the target domain~\cite{fewshot}. This approach can improve segmentation accuracy but requires labeled data from the target domain, thereby compromising the zero-shot nature of the task. 
Advanced methods like SAA+~\cite{SAA} and PromptAD~\cite{promptad} use domain-specific knowledge and context to create precise and pre-defined prompts. Descriptive prompts, such as “a crack in a metal surface,” enhance segmentation accuracy. However, manual curation limits generalization and increases reliance on domain expertise.

%% file: 3.methodology.tex
\section{Problem Definition} \label{sec:definition}
This paper addresses the problem of segmenting unseen anomalous regions in images by predicting an anomaly score \( Score_a \in [0, 1] \) for each pixel, where 0 indicates a normal region and 1 indicates a highly anomalous region. Unlike traditional methods that rely on pre-defined categories, ZSAS approach generalizes to previously unseen anomalies. 

Given a set of images \(I = \{ I_n, I_a \}\), where \(I_n\) are normal images and \(I_a\) are anomalous images, the goal is to generate segmentation masks \(M_a\) and an anomaly score \(Score_a\) for each anomalous image. The segmentation masks \(M_a = \{m_1, m_2, \ldots, m_j\}\) are generated using SAM~\cite{SAM}, and the anomaly score is calculated as \(Score_a = \sum_{i=1}^{j} s_i \cdot m_i\). The confidence scores \(s_i\) are obtained via Grounding DINO~\cite{dino}, which produces bounding boxes \(B = \{b_1, b_2, \ldots, b_k\}\) and corresponding confidence scores \(S = \{s_1, s_2, \ldots, s_k\}\). 

\begin{figure*}[t]
  \centering
  \includegraphics[width=1\linewidth]{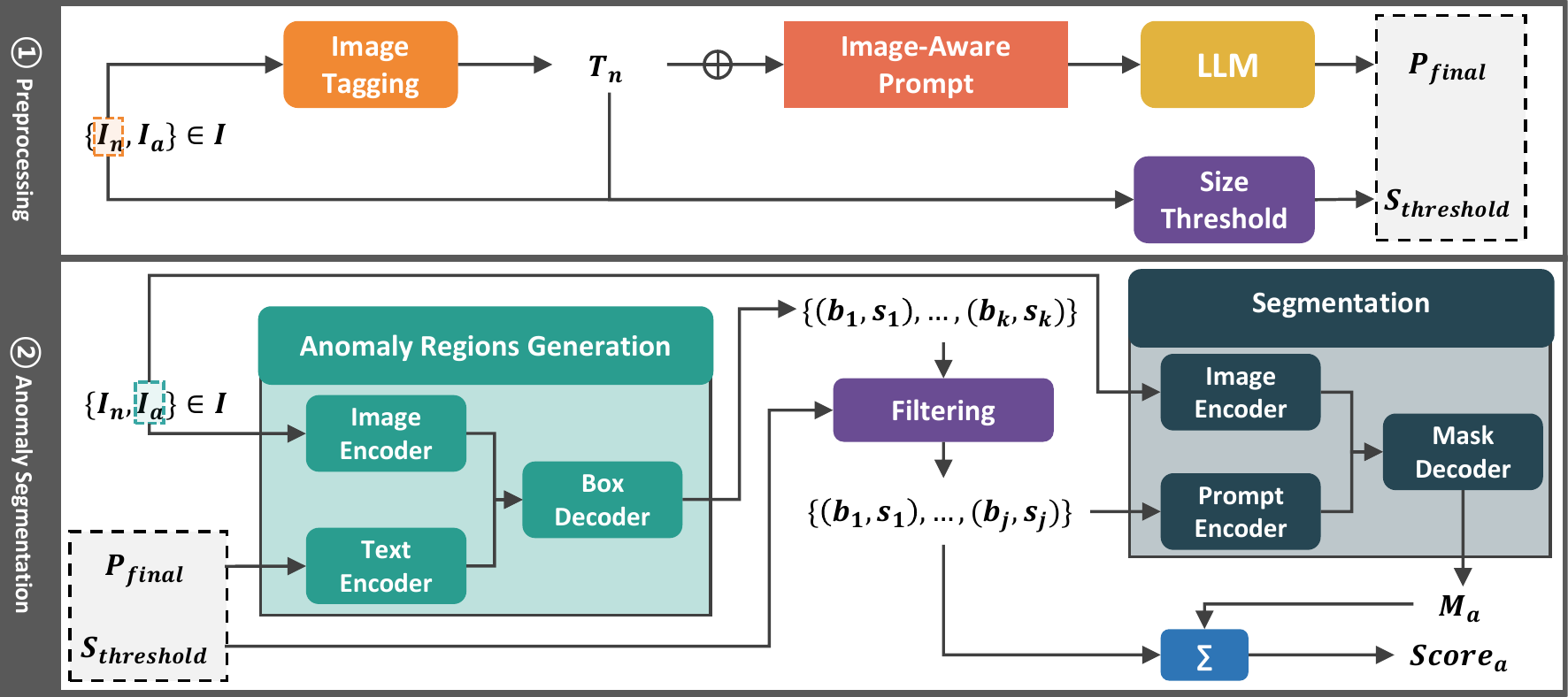}
  \caption{Overview of the proposed IAP-AS framework, which operates in two stages: Preprocessing and Anomaly Segmentation. The Preprocessing stage includes image tagging, size threshold extraction, and LLM-based prompt generation. The Anomaly Segmentation stage involves anomaly region detection, filtering, segmentation, and anomaly score computation.}
  \label{fig:framework}
\end{figure*}

\begin{algorithm}[ht]
\caption{IAP-AS: Image-Aware Prompt Anomaly Segmentation}
\label{alg:IAP-AS}
\begin{algorithmic}[1]

\State \textbf{Stage 1: Preprocessing}
\State \textbf{Input:} Image set \(I = \{ I_n, I_a \}\), Random value \(r \), Image-Aware Prompt (IAP)
\State \textbf{Output:} Final text prompts \( P_{final} \), Size threshold \( S_{\text{threshold}} \)
\For{each normal image \( I_n \) selected randomly with \( r \)}
    \State \( T_n \gets Tagging(I_n) \) \Comment{Step 1-1}
\EndFor
\State \( S_{\text{threshold}} \gets GroundingDINO(I_a, T_n) \) \Comment{Step 1-2}
\State \( P_{final} \gets LLM(\text{IAP} \oplus T_n) \) \Comment{Step 1-3}

\State \textbf{Stage 2: Anomaly Segmentation}
\State \textbf{Input:} Anomaly images \( I_a \), \( P_{final} \), and \( S_{\text{threshold}} \)
\State \textbf{Output:} Segmentation masks \( M_a \), Anomaly score \( Score_a \)
\For{each anomaly image \( I_a \)}
    \State \( B, S \gets GroundingDINO(I_a, P_{final}) \) \Comment{Step 2-1 
    }
    \State \( B_{\text{filtered}}, S_{\text{filtered}} \gets Filtering(B, S, S_{\text{threshold}}) \) \Comment{Step 2-2 
    }
    \State \( M_a \gets SAM(I_a, B_{\text{filtered}}) \) \Comment{Step 2-3 
    }
    \State \( Score_a \gets \sum_{i=1}^{j} s_i \cdot m_i \) \Comment{Step 2-4}
\EndFor
\end{algorithmic}
\end{algorithm}

\section{IAP-AS: Image-Aware Anomaly Segmentation} \label{sec:methodology}
We propose IAP-AS, an enhanced ZSAS method that uses an Image-Aware Prompt (IAP) for flexible segmentation. 
IAP-AS builds on the Grounding DINO with SAM pipeline~\cite{text2seg,SAA} by dynamically generating context-aware prompts, enabling adaptive prompt generation, precise segmentation, and effective anomaly scoring for improved performance.
As shown in Fig.~\ref{fig:framework} and Alg.~\ref{alg:IAP-AS}, IAP-AS operates in two stages: \textit{Preprocessing} (Sec.~\ref{subsec:preprocessing}) and \textit{Anomaly Segmentation} (Sec.~\ref{subsec:AS}).

\begin{enumerate}
    \item In \textit{Preprocessing} stage, three key steps are performed. Step 1-1 extracts object tags from normal images \( I_n \) to identify reference patterns. Step 1-2 calculates a size threshold \( S_{\text{threshold}} \) to filter irrelevant regions. Step 1-3 generates a final text prompt \( P_{final} \) using an LLM to create adaptive context-aware prompts.
    \item In \textit{Anomaly Segmentation} stage, four key steps are executed. Step 2-1 identifies anomaly regions using Grounding DINO. Step 2-2 filters out irrelevant regions based on the size threshold \( S_{\text{threshold}} \). Step 2-3 generates segmentation masks \( M_a \) using SAM for precise anomaly localization. Step 2-4 computes the anomaly score \( Score_a \) by aggregating segmentation masks $M_a$ and confidence scores $S_{\text{filtered}}$.
\end{enumerate}

\subsection{Stage 1: Preprocessing} \label{subsec:preprocessing}
\subsubsection{Step 1-1: Image Tagging}
Semantic labels are assigned to objects in normal images using the Recognize Anything Model (RAM)~\cite{RAM}. The process is defined as \( T_n = \text{Tagging}(I_n) \), where \( T_n \) denotes the set of object tags extracted from normal images \( I_n \). Normal images are selected using a random value \( r \), forming the tag set \( T_n \), and irrelevant tags like "crack" or "scratch" are avoided. These tags are then used for size threshold extraction and LLM-based prompting in subsequent steps, supporting more effective anomaly segmentation. 
Image tagging improves overall performance, as evidenced by ablation studies in Sec.~\ref{sec:Ablation}.
\subsubsection{Step 1-2: Size Threshold}
The size threshold \( S_{\text{threshold}} \) is defined as a specific value to facilitate improved segmentation accuracy. This threshold is calculated as the maximum size of detected regions in the input image \( I_a \), using Grounding DINO and the object tag set \( T_n \). The threshold is then applied in the subsequent anomaly segmentation stage (Sec.~\ref{subsec:AS}), where it filters out smaller regions and contributes to the improvement of segmentation accuracy.
\subsubsection{Step 1-3: LLM Prompting}
This step uses an LLM to generate dynamic, context-aware text prompts for text-guided ZSAS. Unlike fixed prompts (e.g., "defect," "abnormal"), our approach tailors prompts to each dataset's characteristics. The model employs LLaMA-3-8B~\cite{Llama} to generate context-specific tags \(P_{adj}\), enhancing adaptability and segmentation accuracy by tailoring prompts to each dataset. Ablation studies 
validate its effectiveness in anomaly detection and segmentation.
This process uses IAP to generate anomaly-related tags from the object tag set \(T_n\). IAP and \(T_n\) are merged and input into an LLM to generate adjective clauses \(P_{adj}\), which help create context-aware, dataset-specific prompts for text-guided ZSAS. The process is defined as:
\begin{equation}
    P_{adj} = LLM(IAP \oplus T_n).
\label{eq:llm}
\end{equation}
In Eq.~\eqref{eq:llm}, \(\oplus\) denotes the merge operation and \( P_{adj} \) represents the adjective clauses generated by \textit{LLM}. These clauses help create context-aware prompts, allowing the model to generate dataset-specific prompts for text-guided ZSAS.
The generated adjective clauses \( P_{adj} \) are combined with fixed prompts ("abnormal", "defect") and the object tag set \( T_n \) from image tagging to create the final prompt \( P_{final} \). This process removes redundancy and allows for flexible, dataset-specific prompts, improving anomaly segmentation accuracy.

\subsection{Stage 2: Anomaly Segmentation} \label{subsec:AS}
\subsubsection{Step 2-1: Anomaly Regions Generation}
Grounding DINO~\cite{dino} is used to localize potential anomalous regions in the input image \(I_a\). The image encoder extracts image features \(I_{df} = \text{ImageEncoder}(I_a)\), while the text encoder generates text features \(P_{df} = \text{TextEncoder}(P_{final})\) from the text prompts. These features are combined and passed to a Box Decoder, which produces a set of bounding boxes \(B = \{b_1, b_2, \ldots, b_k\}\) and corresponding confidence scores \(S = \{s_1, s_2, \ldots, s_k\}\) as:  
\begin{equation}
\label{eq:mask}
    \{(b_i, s_i) \mid i = 1, 2, \ldots, k\} = \text{BoxDecoder}(I_{df}, P_{df}).
\end{equation}
Each bounding box \(b_i\) and confidence score \(s_i\) represents a potential anomaly region, where higher confidence scores indicate a higher likelihood of containing an anomaly. The resulting bounding boxes \(B\) are then passed to the filtering step to retain the most relevant anomaly regions.
\subsubsection{Step 2-2: Filtering}
The filtering step improves anomaly detection by excluding regions larger than a predefined size threshold \(S_{threshold}\) and retaining only smaller regions. This process is defined as:
\begin{equation}
    \begin{array}{c}
        ({B}_{\text{filtered}}, {S}_{\text{filtered}}) = \{(b_i, s_i) \mid i \in \mathcal{J}\} \\
        \text{where} \quad \mathcal{J} = \{ i \mid b_i < S_{\text{threshold}}, \, i \in \{1, 2, \ldots, k\} \}
    \end{array}
    \label{eq:filtering}
\end{equation}
Here, \({B}_{filtered}\) and \({S}_{filtered}\) denote the filtered bounding boxes and their confidence scores, where \(j \leq k\). The filtered regions are passed to the segmentation step, ensuring that only relevant regions are further processed. This size-based filtering has been shown to enhance overall performance, as demonstrated by the ablation studies in Sec.~\ref{sec:Ablation}.

\subsubsection{Step 2-3: Segmentation}
After the filtering step, the bounding boxes \({B}_{filtered}\) are passed to SAM~\cite{SAM} to generate detailed segmentation masks for the anomalous regions. The image encoder extracts image features \(I_{sf} = \text{ImageEncoder}(I_a)\), while the prompt encoder processes \({B}_{filtered}\) to generate prompt features \({B}_{sf} = \text{PromptEncoder}({B}_{filtered})\). These features are combined and processed by the mask decoder to produce refined segmentation masks \(M_a\) defined as:
\begin{equation}
    M_a = \{m_1, m_2, \ldots, m_j\} = \text{MaskDecoder}(I_{sf}, {B}_{sf}).
    \label{eq:segmentation}
\end{equation}
Each segmentation mask \(m_i\) corresponds to the bounding box \(b_i\) in \({B}_{filtered}\), providing a fine-grained analysis of anomalous regions, which results in the final anomaly mask.

\subsubsection{Step 2-4: Anomaly Score}
The anomaly segmentation score \(Score_a\) is calculated by aggregating segmentation masks \(M_a\) with their corresponding confidence scores \({S}_{filtered}\), as given by \(Score_a = \sum_{i=1}^{j} s_i \cdot m_i\).
This weighted approach emphasizes regions with higher anomaly confidence while reducing the impact of less likely areas. The resulting anomaly score \(Score_a\) provides a comprehensive representation of anomalous regions, enhancing localization and prioritization for more accurate and reliable segmentation.

%% file: 4.experiments.tex
\section{Experiments}
\label{sec:experiments}
\subsection{Implementation Details}
\subsubsection{Datasets.} Our experiment comprehensively evaluates the model using a total of seven major industrial datasets, which are systematically divided into object and texture datasets: MVTec AD~\cite{bergmann2019mvtec}, MPDD~\cite{mpdd}, BTAD~\cite{btad}, KSDD1~\cite{Tabernik2019JIM}, MTD~\cite{huang2020surface}, DTD-Synthetic~\cite{dtd}, and DAGM2007~\cite{dagm}, each of which is widely recognized for its high-resolution, industrial-rich images that are particularly well-suited for effective anomaly segmentation tasks.

\subsubsection{Evaluation Metrics.} For anomaly segmentation, we evaluate performance using two metrics: Average Precision (AP) and the maximum F1 score (F1-max). Calculated as the area under the precision-recall curve, AP measures the trade-offs between precision and recall, while F1-max represents the highest F1 score across all thresholds. Both metrics are reported as percentages for comparison.

\subsubsection{Parameters.} The experimental settings used box and text thresholds of 0.2 for object datasets and 0.1 for texture datasets, accurately reflecting the need for finer anomaly detection. An IoU threshold of 0.5 effectively removed overlapping detection boxes, a size threshold of 0.8 filtered out unusually sized bounding boxes, and a fixed random seed of 111 ensured experimental reproducibility.

\subsubsection{Comparison.} For model comparison, official implementations were used where available. WinCLIP~\cite{WinCLIP} utilized the implementation from~\cite{SAA}. Text prompts were customized to match each model's characteristics. CLIPseg~\cite{ClipSeg} and EVF-SAM~\cite{evf} use fixed prompts as object detection models, while WinCLIP, SAA+~\cite{SAA}, and SDP~\cite{clipad} employ dataset-specific Few-Shot prompts combined with object names.

\subsection{Quantitative Results}
\input{tb.ZSAS-evaluation}
\begin{table*}[t]
\centering
\footnotesize
\setlength{\abovecaptionskip}{5pt}
\setlength{\tabcolsep}{5pt}
\renewcommand{\arraystretch}{1.2}
\resizebox{\textwidth}{!}{
\begin{tabular}{@{}c|c|c@{}}
  \toprule
  Image & Object Tag Set (\(T_n\)) & Text Prompts (\(P_{final}\)) \\
  \midrule
  Carpet & [`cloth fabric gray material pattern texture'] & [discoloration, fray, rip, bubble, stain, burn, \ldots]\\
  Leather & [`brown fabric leather material skin \ldots'] & [cut, hole, split, blemish, tear, mark, \ldots]\\
  Grid & [`floor grid', `grid net pattern texture'] & [disruption, disturbance, aberration, stain, \ldots]\\
  Tile & [`granite gray marble stone surface \ldots'] & [flake, blemish, mark, flaw, pit, cavity, stain, \ldots]\\
  Wood & [`hardwood floor wood \ldots'] & [insect, rot, splinter, stain, crack, decay, \ldots] \\
  \bottomrule
\end{tabular}
}
\caption{Example of IAP-AS preprocessing on MVTec-AD texture dataset.}
\label{tab:preprocessing}
\end{table*}
We evaluated IAP-AS against several state-of-the-art text-guided ZSAS methods that do not require training. The results are detailed in Table~\ref{tab:results}. On all datasets, IAP-AS demonstrates superior performance by achieving high AP and F1-max scores. The AP score represents the ability to detect true anomalies with minimal false positives, while the F1-max score reflects the balance between recall and precision for accurately identifying defect locations. 
Notably, on object datasets such as MVTec-AD~\cite{bergmann2019mvtec} and texture datasets like KSDD1~\cite{Tabernik2019JIM}, MTD~\cite{huang2020surface}, and DAGM2007~\cite{dagm}, IAP-AS outperforms existing models across all metrics, delivering balanced and robust results. Its exceptional ability to handle the complex characteristics of industrial objects and textures underscores its advantages in practical applications. On object datasets such as MPDD~\cite{mpdd} and BTAD~\cite{btad}, IAP-AS achieves up to a 10.99\% improvement in F1-max, ranking first, and secures second place in AP. Furthermore, on texture datasets like DTD-Synthetic~\cite{dtd}, IAP-AS demonstrates excellent performance, ranking second by a narrow margin behind SAA+. These results underscore IAP-AS's strong generalization capabilities for diverse anomalies in industrial datasets and its adaptability to ensure reliable performance in real-world scenarios.

\subsection{Qualitative Results}
\begin{figure*}[t]
  \centering
\includegraphics[width=1.0\linewidth]{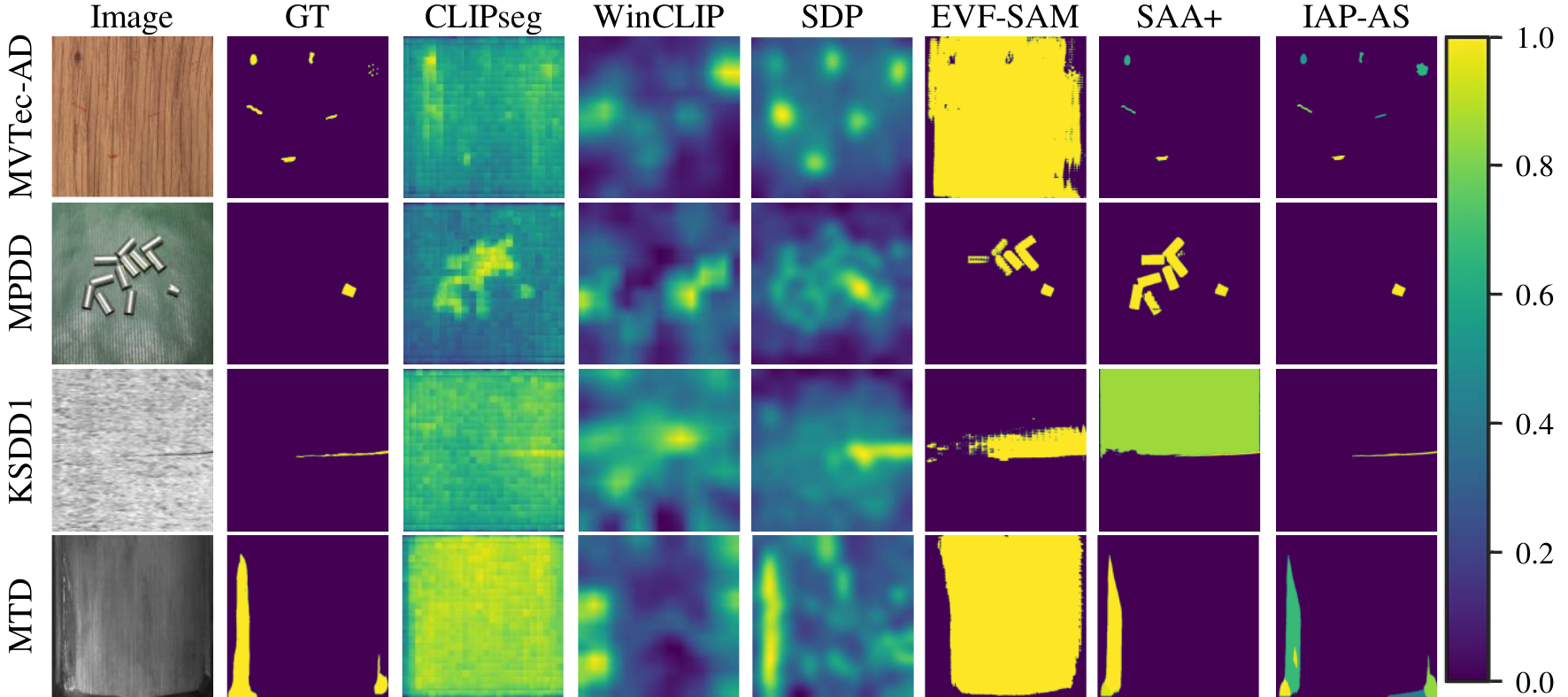}
\caption{Visual comparison of IAP-AS and other models across four datasets, highlighting differences in anomaly segmentation results.}
\label{fig:fig_example}
\end{figure*}
A qualitative evaluation was conducted to validate the anomaly segmentation performance of IAP-AS using industrial datasets. Two object datasets (MVTec-AD~\cite{bergmann2019mvtec} and MPDD~\cite{mpdd}) and two texture datasets (KSDD1~\cite{Tabernik2019JIM} and MTD~\cite{huang2020surface}) were selected for the assessment. Fig.~\ref{fig:fig_example} provides a visual comparison of the original images, ground truth anomaly masks, and the anomaly score maps produced by both the comparison models and IAP-AS.
IAP-AS excels in segmenting anomalies in object datasets, accurately detecting defects even in complex scenarios with multiple flaws or coexisting normal and abnormal objects. It utilizes dynamically adaptive image recognition prompts to detect anomalies of varying sizes and locations with high accuracy, aligning closely with ground truth masks. On texture datasets, IAP-AS not only accurately segments large anomaly regions but also captures fine defects with exceptional precision. As a SAM-based model, it efficiently handles fine and coarse anomalies across diverse sizes and shapes.

The power of IAP-AS comes from its ability to generate high-quality IAPs. Fig.~\ref{fig:Prompting-example} provides a visual representation of an example of IAP generation, and Table~\ref{tab:preprocessing} provides a more detailed example. The table shows how simple tagging information \( T_n \) from the MVTec-AD texture dataset is expanded into domain-specific prompts \( P_{final} \) enriched with contextual details using LLM. This shows that LLM enables precise anomaly detection by dynamically generating domain-appropriate adjectives, and the \( P_{final} \) generated in this way plays a key role in helping IAP-AS achieve high anomaly detection accuracy. These results show that IAP-AS ensures accurate and reliable outlier segmentation in challenging industrial scenarios requiring detailed detection. It excels in handling multiple outliers or complex patterns, as demonstrated on the MVTec-AD dataset, which features diverse anomalies and intricate defect patterns. This highlights its industrial potential and surpasses existing methods in such challenging scenarios.

\subsection{Ablation Study} \label{sec:Ablation}
\input{tb.ZSAS-ablation}

An ablation study was conducted on the MVTec-AD dataset to assess the contribution of key components in the IAP-AS model. The results are detailed in Table~\ref{tab:ablation}. The key steps (Sec.~\ref{subsec:preprocessing} and~\ref{subsec:AS}) include image tagging to extract an object tag set \(T_n\) (Step 1-1), dynamic prompting to generate the final text prompt \(P_{final}\) (Step 1-3), and filtering to limit anomaly size using the threshold \(S_{\text{threshold}}\) (Step 2-2).
To evaluate the impact of key components, we designed six model variants by selectively including or excluding specific components while keeping the hyperparameters constant. The impact of each component was assessed using AP and F1-max metrics, showing that progressively adding components improves anomaly detection and segmentation performance.
The baseline model achieved an AP of 12.6\% and an F1-max of 24.73\%. Image tagging alone showed limited improvement, but when combined with LLM prompting, the performance improved significantly, highlighting the crucial role of domain-specific prompt generation in anomaly detection. Finally, by adding filtering, the size of anomaly regions was effectively constrained, leading to further improvements in detection accuracy. These results demonstrate that all components work complementarily, creating a synergistic effect that enhances overall performance.
Integrating all three components achieved the best results, with an AP of 31.43\% and an F1-max of 40.13\%, showing a 9.18\% improvement over the baseline. This confirms the superior performance of the fully integrated IAP-AS model.

%% file: tb.ZSAS-evaluation.tex
\begin{table*}[t]
\centering
\footnotesize
\setlength{\abovecaptionskip}{5pt}
\setlength{\tabcolsep}{5pt}
\renewcommand{\arraystretch}{1.11}
\resizebox{\textwidth}{!}{
\begin{tabular}{cc|c|ccc|ccc}
\hline
\multicolumn{2}{c|}{\multirow{2}{*}{Datasets}}  & \multirow{2}{*}{Metrics} & \multicolumn{3}{c|}{CLIP-based Models}  & \multicolumn{3}{c}{SAM-based Models} \\ \cline{4-9} 
\multicolumn{2}{c|}{} &   & \multicolumn{1}{c|}{CLIPseg} & \multicolumn{1}{c|}{WinCLIP}     & SDP            & \multicolumn{1}{c|}{SAA+}           & \multicolumn{1}{c|}{EVF-SAM}     & IAP-AS          \\ \cline{3-9} 
\hline

\multicolumn{1}{c|}{\multirow{6}{*}{Object}}  & \multirow{2}{*}{MVTec-AD}      & AP                     & \multicolumn{1}{c|}{20.08}   & \multicolumn{1}{c|}{17.77}       & 30.60          & \multicolumn{1}{c|}{\underline{28.83}}    & \multicolumn{1}{c|}{6.67}        & \textbf{31.43} \\
\multicolumn{1}{c|}{}                         &                                & F1-max                 & \multicolumn{1}{c|}{24.06}   & \multicolumn{1}{c|}{24.76}       & 1.80           & \multicolumn{1}{c|}{\underline{37.70}}   & \multicolumn{1}{c|}{13.01}       & \textbf{40.13} \\ \cline{2-9} 
\multicolumn{1}{c|}{}                         & \multirow{2}{*}{MPDD}          & AP                     & \multicolumn{1}{c|}{11.27}   & \multicolumn{1}{c|}{13.36}       & \textbf{14.70} & \multicolumn{1}{c|}{8.68}           & \multicolumn{1}{c|}{12.55}       & {\underline{13.38}}    \\
\multicolumn{1}{c|}{}                         &                                & F1-max                 & \multicolumn{1}{c|}{14.75}   & \multicolumn{1}{c|}{14.73}       & 2.30           & \multicolumn{1}{c|}{17.27}          & \multicolumn{1}{c|}{\underline{21.89}} & \textbf{24.01} \\ \cline{2-9} 
\multicolumn{1}{c|}{}                         & \multirow{2}{*}{BTAD}          & AP                     & \multicolumn{1}{c|}{5.76}    & \multicolumn{1}{c|}{8.70}        & \textbf{21.80} & \multicolumn{1}{c|}{5.96}           & \multicolumn{1}{c|}{4.82}        & {\underline{16.97}}    \\
\multicolumn{1}{c|}{}                         &                                & F1-max                 & \multicolumn{1}{c|}{12.25}   & \multicolumn{1}{c|}{\underline{14.63}} & 2.60           & \multicolumn{1}{c|}{12.35}          & \multicolumn{1}{c|}{9.33}        & \textbf{25.62} \\ \hline
\multicolumn{1}{c|}{\multirow{8}{*}{Texture}} & \multirow{2}{*}{KSDD1}         & AP                     & \multicolumn{1}{c|}{4.12}    & \multicolumn{1}{c|}{1.54}        & 9.30           & \multicolumn{1}{c|}{\underline{12.18}} & \multicolumn{1}{c|}{0.09}        & {\textbf{13.91}}     \\
\multicolumn{1}{c|}{}                         &                                & F1-max                 & \multicolumn{1}{c|}{8.85}    & \multicolumn{1}{c|}{3.73}        & 6.50           & \multicolumn{1}{c|}{\underline{17.90}} & \multicolumn{1}{c|}{0.32}        & {\textbf{22.63}}    \\ \cline{2-9} 
\multicolumn{1}{c|}{}                         & \multirow{2}{*}{MTD}           & AP                     & \multicolumn{1}{c|}{6.75}    & \multicolumn{1}{c|}{3.27}        & 6.90           & \multicolumn{1}{c|}{\underline{27.39}}    & \multicolumn{1}{c|}{2.38}        & \textbf{27.77} \\
\multicolumn{1}{c|}{}                         &                                & F1-max                 & \multicolumn{1}{c|}{9.43}    & \multicolumn{1}{c|}{8.48}        & 14.00          & \multicolumn{1}{c|}{\underline{37.27}}    & \multicolumn{1}{c|}{5.09}        & \textbf{41.63} \\ \cline{2-9} 
\multicolumn{1}{c|}{}                         & \multirow{2}{*}{DTD-Synthetic} & AP                     & \multicolumn{1}{c|}{19.75}   & \multicolumn{1}{c|}{9.96}        & 37.70          & \multicolumn{1}{c|}{\textbf{50.10}} & \multicolumn{1}{c|}{4.54}        & {\underline{49.23}}    \\
\multicolumn{1}{c|}{}                         &                                & F1-max                 & \multicolumn{1}{c|}{23.77}   & \multicolumn{1}{c|}{16.60}       & 16.70          & \multicolumn{1}{c|}{\textbf{62.29}} & \multicolumn{1}{c|}{9.32}        & {\underline{60.33}}   \\ \cline{2-9} 
\multicolumn{1}{c|}{}                         & \multirow{2}{*}{DAGM2007}      & AP                     & \multicolumn{1}{c|}{20.94}   & \multicolumn{1}{c|}{17.05}       & 14.20          & \multicolumn{1}{c|}{\underline{22.72}}          & \multicolumn{1}{c|}{2.92}        & \textbf{24.03} \\
\multicolumn{1}{c|}{}                         &                                & F1-max                 & \multicolumn{1}{c|}{25.82}   & \multicolumn{1}{c|}{25.56}       & 6.70           & \multicolumn{1}{c|}{\underline{33.82}}          & \multicolumn{1}{c|}{5.62}        & \textbf{34.68} \\ \hline
\end{tabular}
}
\caption{Performance comparison of trainig-free text-guided zero-shot anomaly segmentation models  on seven industrial datasets, highlighting AP and F1-max metrics.}
\label{tab:results}
\end{table*}

%% file: tb.ZSAS-ablation.tex
\begin{table}[t]
\centering
\footnotesize
\setlength{\abovecaptionskip}{5pt}
\setlength{\tabcolsep}{5pt}
\renewcommand{\arraystretch}{1.1}
\begin{tabular}{ccc|cc}
  \toprule
  Step 1-1 & Step 1-3   & Step 2-2 & AP & F1-max \\
  \midrule
  X  & X  & X  & 12.67              & 24.73\\
  O  & X  & X  & 17.49              & 28.21\\
  O  & O  & X  & 20.14              & 33.08\\
  X  & O  & X  & 21.25              & 33.66\\
  X  & O  & O  & \underline{21.84}              & \underline{34.06}\\
  O  & O  & O  & \textbf{31.43}     & \textbf{40.13}\\
  \bottomrule
\end{tabular}
\caption{Ablation study of IAP-AS on the MVTec-AD dataset, showing the impact of Image Tagging (Step 1-1), LLM Prompting (Step 1-3), and Filtering (Step 2-2).}
\label{tab:ablation}
\end{table}

%% file: 5.conclusions.tex
\section{Conclusion}
\label{sec:conclusions}
In this paper, we introduce Image-Aware Prompt Anomaly Segmentation (IAP-AS), a novel approach that leverages LLM~\cite{Llama} to dynamically generate image-aware prompts and use Grounding DINO~\cite{dino} and SAM~\cite{SAM} for segmentation to overcome the limitations of text-based ZSAS that rely on fixed, predefined prompts and domain-specific knowledge. Experimental results demonstrate that IAP-AS achieves high precision in anomaly segmentation with up to a 10\% improvement in F1-max compared to CLIP-based methods (e.g., CLIPSeg, WinCLIP, SDP) and SAM-based models (e.g., SAA+, EVF-SAM), while supporting cross-domain generalization without retraining. We release our code to support future research and broader applicability of IAP-AS. In the future, we plan to further refine the design of image recognition prompts optimized for anomaly tag extraction to enable wider applicability across diverse environments. Additionally, we aim to enhance the practical applicability of the proposed method by exploring optimization strategies to ensure high accuracy and reliability in complex real-world scenarios.